\documentclass[conference]{IEEEtran}
\IEEEoverridecommandlockouts
\usepackage{cite}
\usepackage{amsmath,amssymb,amsfonts}
\usepackage{algorithmic}
\usepackage{graphicx}
\usepackage{textcomp}
\usepackage{xcolor}
\usepackage{amsmath}
\usepackage{tabularx}
\usepackage{multirow}
\usepackage{rotating}
\usepackage{algorithm,algorithmic}
\def\BibTeX{{\rm B\kern-.05em{\sc i\kern-.025em b}\kern-.08em
    T\kern-.1667em\lower.7ex\hbox{E}\kern-.125emX}}

\newcolumntype{L}{>{\small\raggedright\let\newline\\\arraybackslash\hspace{0pt}}p{0.7cm}}
\newcolumntype{R}{>{\raggedright\let\newline\\\arraybackslash\hspace{0pt}}X}
\newcolumntype{M}{>{\raggedleft\let\newline\\\arraybackslash\hspace{0pt}}p{1.8cm}}
\newcolumntype{S}{>{\raggedright\let\newline\\\arraybackslash\hspace{0pt}}p{1cm}}

\begin{document}

\title{On the improvement of model-predictive controllers.}

\author{\IEEEauthorblockN{1\textsuperscript{st} Leander J. Féret}
\IEEEauthorblockA{\textit{Think Tank} \\
\textit{JUMO GmbH \& Co. KG}\\
Fulda, Germany \\
leander.feret@jumo.net}
\and
\IEEEauthorblockN{2\textsuperscript{nd} Alexander Gepperth}
\IEEEauthorblockA{\textit{Applied Computer Science Department} \\
\textit{Fulda University of Applied Sciences}\\
Fulda, Germany \\
0000-0003-2216-7808}
\and
\IEEEauthorblockN{3\textsuperscript{rd} Steven Lambeck}
\IEEEauthorblockA{Electrical Engineering Department}
\textit{Fulda University of Applied Sciences}\\
Fulda, Germany \\
steven.lambeck@et.hs-fulda.de
}

\maketitle

\begin{abstract}
This article investigates synthetic model-predictive control (MPC) problems to demonstrate that an increased precision of the internal prediction model (PM) automatially entails an improvement of the controller as a whole. In contrast to reinforcement learning (RL), MPC uses the PM to predict subsequent states of the controlled system (CS), instead of directly recommending suitable actions. To assess how the precision of the PM translates into the quality of the model-predictive controller, we compare a DNN-based PM to the optimal baseline PM for three well-known control problems of varying complexity. The baseline PM achieves perfect accuracy by accessing the simulation of the CS itself. Based on the obtained results, we argue that an improvement of the PM will always improve the controller as a whole, without considering the impact of other components such as action selection (which, in this article, relies on evolutionary optimization).
\end{abstract}


\section{Introduction}
\label{chap:introduction}

The use of machine learning (ML) is widespread in different scientific and industrial use-cases. One of these areas is control science, in which data-driven ideas have long been prevalent, for example in data-driven system identification \cite{b5}, \cite{b6}.

Even some of the best known use-cases of ML have their roots in control science, as many will associate ML with self-trained robots learning to walk \cite{b9}, \cite{b10}, \cite{b11} or fully automated industry productions \cite{b12}. Therefore, ML is not a new topic in the field of control science.

In general, when systems are difficult to control with a traditional PID-controller or when the creation of a mathematical model for an MPC \cite{b1} or the training of reinforcement learning (RL) is problematic, the use of data-driven system-identification is a common option.

With the recent advances in neural networks, many researchers are now working on using neural networks as PMs for ML-MPCs \cite{b2}, \cite{b3}, \cite{b7}, \cite{b8}.

This article provides support for the claim that the quality of ML-based model-predictive control mainly (ML-MPC) depends on the prediction accuracy of its PM. From this, it is argued, that current and upcoming research on this topic, can be strictly focused on the core components of an MPC without showing the improvements of the whole in practical setups.

\begin{figure}[htbp]
	\centerline{\includegraphics[width=8.2cm]{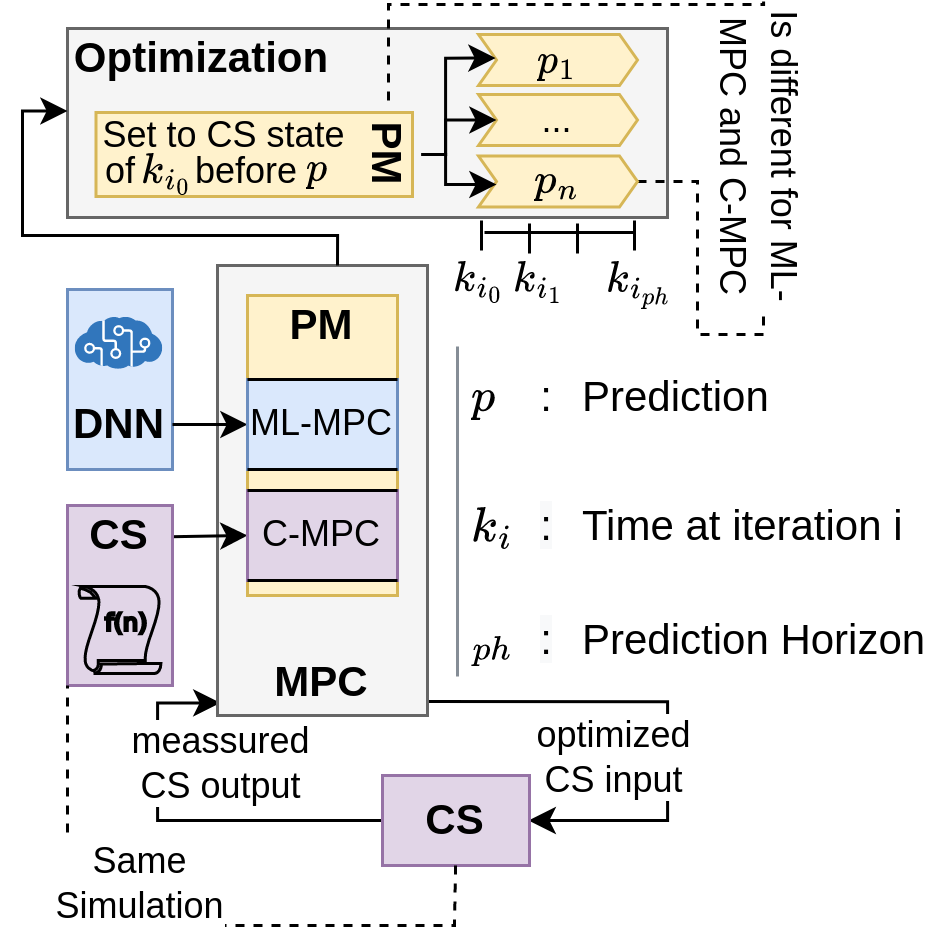}}
	\caption{A simplified control loop for an MPC and its core components.}
	\label{fig:mpc}
\end{figure}

To provide a strong base for this argument, we employ a basic ML-MPC architecture as shown in Fig. \ref{fig:mpc}.

This allows us to eliminate any possible architectural reasons for the loss of control quality, except for the differences resulting from the different PMs.

Consequently, only two points of interest remain, which are further discussed in section \ref{cha:model_predictive_control} and \ref{cha:discussion}.

For an overview of the experimental setup, all used core parts of both ML-MPC and comparison MPC (C-MPC), are discussed in section \ref{cha:simulation}, \ref{sec:data_preperation}, \ref{sec:prediction_models} and \ref{cha:model_predictive_control}. Results are presented in \ref{sec:comparison}.

Within \ref{cha:discussion} and \ref{cha:conclusion} it is stated that the prediction accuracy is the main reason for any given control discrepancy between the ML-MPC and the C-MPC.

\subsection{Related Work}

As this article shows, ML-MPC can be split into data-driven system-identification, non-linear optimization and model-predictive control. Thus, every contribution revolving around these topics is directly related to the building of an ML-MPC.

For example, data-driven system-identification is discussed in \cite{b13}, \cite{b14}, and non-linear optimization in \cite{b15}, and within data-science in \cite{b16}, \cite{b19}, \cite{b20}.

MPC is a stable technology within control engineering, while ML-MPC is discussed in several domains \cite{b18}, \cite{b21}. It is often analyzed regarding different prediction models \cite{b21}, \cite{b22}, \cite{b23}. A prominent discussion of this aspect is given in \cite{b17}, which advocates the usage of an LSTM system customized with dropout layers to better predict noisy data, while preventing overfitting and using it for model-predictive control. It is shown that dropout LSTM and co-teaching LSTM give better results on non-Gaussian noisy data than the standard LSTM network does. While the core topic is data-driven system identification, the trained PMs are still used in the context of ML-MPC, and applied to a real chemical system. This is technically unnecessary, as, as shown in this article, any proven better network prediction is also directly improving any ML-MPC. The experimental setup could just be used to show how the prediction improved, without the need for an ML-MPC. This would reduce information overhead and complexity.

\subsection{Contribution}

First of all, this article aims to give a good overview of the basic architecture of a simple ML-MPC and C-MPC and its components.

More importantly, it suggests that the problem of finding good ML-MPCs can be split into sub-problems for future research, while giving an experimental evidence to justify this decision.

This way, the next iteration of work on ML-MPC can focus on the issues ahead, like the research on better system-identification or the work on non-linear optimization algorithms. This will reduce the overhead of work and complexity, and will hopefully result in cleaner and more minimalistic future problem formulations within this field of research.

\section{Simulations}
\label{cha:simulation}

In classical control theory, non-linear systems are often controlled by linearizing around fixed points and then applying solutions from linear control theory. This type of linearization can become quite difficult where CS are strongly non-linear. Therefore, ML may be able to generate easier and even better control solutions. To generate training data and targets for the DNN, the following three simulations of non-linear physical dynamic systems with varying complexity are used. They will then be used as the CS, as well as the the PM for C-MPC.

\subsection{Pendulum}

\begin{figure}[htbp]
	\centerline{\includegraphics[width=9cm]{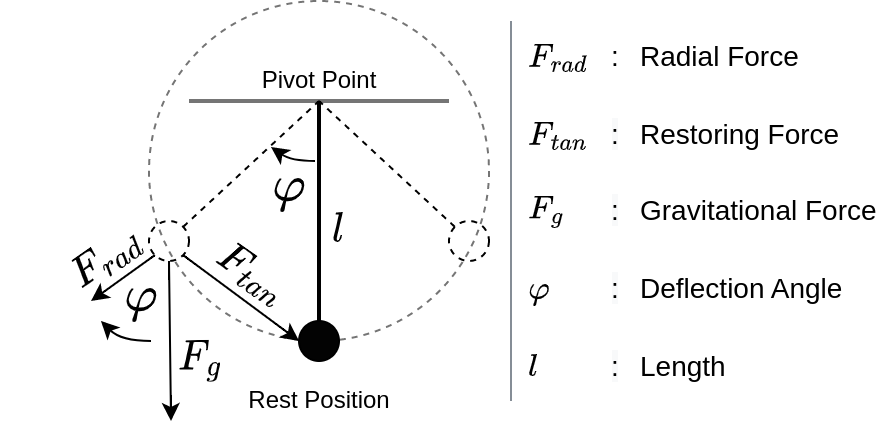}}
	\caption{Idealized, mathematical pendulum.}
	\label{fig_pendulum}
\end{figure}

The first simulation is an idealized, mathematical model of a pendulum, defined as a bob of constant mass $\mathit{m [kg]}$ on a rigid, weightless rod of a constant length $\mathit{l [m]}$, which is fixed to a pivot point with one degree of freedom as shown in Fig. \ref{fig_pendulum}.

Here, the gravitational force is defined as $\mathit{F_{g} = m \cdot g}$, where $\mathit{g}$ is the acceleration of gravity on earth.

The deflection angle $\varphi [^\circ]$ is defined as the angle between the rod and the rest position of the pendulum, which is the position perpendicular below the pivot point.

If the pendulum is not at the rest position, a restoring force $\mathit{F_{tan}}$ will act on it, pointing towards the rest position. The restoring force increases proportional to the deflection angle $\varphi$. As the force vector is pointing toward the rest position, (\ref{eq_restoring_force}) has a negative sign.

\begin{equation}
	F_{tan}(t) = - m \cdot g \cdot \sin(\varphi(t))
	\label{eq_restoring_force}
\end{equation}

Since the pendulum can only move in a circle around the pivot point, the acceleration is tangential, also called a tangential acceleration $\mathit{a_{tan}} [^\circ \cdot s^{-2}]$, which is defined as the angular acceleration $\ddot{\varphi} [^\circ \cdot s^{-2}]$ multiplied by the length $\mathit{l}$ of the pendulum.

\begin{equation}
	a_{tan}(t) = l \cdot \ddot{\varphi}
	\label{eq_tangential_acceleration}
\end{equation}

According to Newton's second law, the equation of motion for the restoring force can be written as:

\begin{equation}
	F_{tan}(t) = m \cdot a_{tan}(t)
	\label{eq_restoring_force_with_acceleration}
\end{equation}

Using the previous equations, the relationship can be written as:

\begin{equation}
	\ddot{\varphi} + \frac{g}{l} \cdot \sin(\varphi(t)) = 0
	\label{eq_transformed_equated_restoring_force}
\end{equation}

According to Newtons First Law, wherein the net force acting on an idealized point mass is the sum of all involved forces, the next angular velocity $\mathit{\hat{\dot{\varphi}}}$ can be written as:

\begin{equation}
	\label{eq_next_velocity_overview}
	\hat{\dot{\varphi}} = F_{vel} + F_{acc} + F_{in}
\end{equation}

with the velocity force $\mathit{F_{vel}}$, the angular acceleration force $\mathit{F_{acc}}$ and an arbitrary input force $\mathit{F_{in}}$.
		
To calculate the angular velocity force $\mathit{F_{vel}}$, (\ref{eq_velocity_force}) is used, where the current velocity $\mathit{\dot{\varphi}}$ is adjusted in relation to the friction $\mathit{b}$, and the pendulum constants, pointing to the resting position and therefore including a negative sign.

\begin{equation}
	\label{eq_velocity_force}
	F_{vel} = - \frac{b}{m \cdot l^2} \cdot \dot{\varphi}
\end{equation}

The angular acceleration force $F_{acc}$ is the result of the gravitational pull acting on the pendulum towards its rest position as discussed in the previous equations regarding the tangential restoration force $\mathit{F_{tan}}$. From this, the angular acceleration $\mathit{\ddot{\varphi}}$ can be calculated, as shown in (\ref{eq_transformed_equated_restoring_force}) and in (\ref{eq_acceleration_force}) in the context of the acceleration force $F_{acc}$.

\begin{equation}
	\label{eq_acceleration_force}
	F_{acc} = - \frac{g}{l} \cdot \sin(\varphi)
\end{equation}

The external input force $\mathit{F_{in}}$ is an arbitrary force, given by an input value $\mathit{u} [N]$ and the physical values of the pendulum shown in (\ref{eq_input_force}).

\begin{equation}
	\label{eq_input_force}
	F_{in} = \frac{1}{m \cdot l^2} \cdot u
\end{equation}
\\

\subsection{Cartpole}

\begin{figure}[htbp]
	\centerline{\includegraphics[width=9cm]{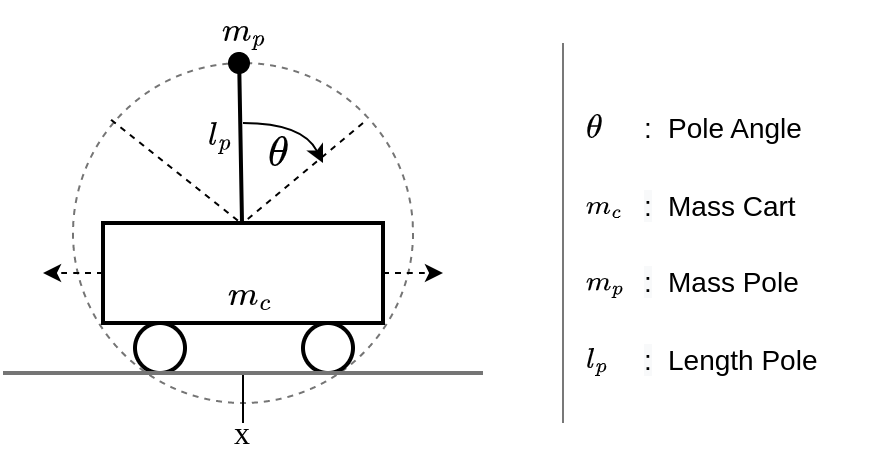}}
	\caption{Abstracted Cartpole.}
	\label{fig_cartpole}
\end{figure}

The second model is the mathematical cartpole, which represents a cart of constant mass $\mathit{m_{c}[kg]}$, that has a pole with a constant mass $\mathit{m_{p}[kg]}$ connected to it. The pole has a constant length $\mathit{l_{p}[m]}$ and a theoretical balanced position orthogonal to the top of the cartpole. The pole angle $\mathit{\theta[^\circ]}$ is defined regarding this balanced position. The cartpole has only one degree of freedom, where its position $\mathit{x} [m]$ is known. This system is shown in Fig. \ref{fig_cartpole}.

Four values represent the state of the system $\mathit{S}$, which are the cart's position $\mathit{x}$, the cart's velocity $\mathit{\dot{x} [m \cdot s^{-1}]}$, the pole angle $\mathit{\theta [^\circ]}$ and the angular velocity of the pole $\mathit{\dot{\theta} [^\circ \cdot s^{-1}]}$.

The simulation of the cartpole is performed in discrete time, and therefore an arbitrary timestep variable $\tau [s]$ is used.

For each time step within the simulation, the next system state must be calculated, while using the previous system state and the physical values of the cartpole. In conclusion, the system state is defined as $S=[x, \dot{x}, \theta, \dot{\theta}]$ and, correspondingly, the next system state as $\hat{S} = [\hat{x}, \hat{\dot{x}}, \hat{\theta}, \hat{\dot{\theta}}]$. Its components can be calculated using the following equations:

\begin{align} \label{eq_next_cartpole_state_overview}
		\hat{x} &= x + \tau \cdot \dot{x}&\\ \nonumber
		\hat{\dot{x}} &= \dot{x} + \tau \cdot \ddot{x}&\\ \nonumber
		\hat{\theta} &= \theta + \tau \cdot \dot{\theta}&\\ \nonumber
		\hat{\dot{\theta}} &= \dot{\theta} + \tau \cdot \ddot{\theta}
\end{align}

As the system state $\mathit{S}$ is monitored, only the cart acceleration $\ddot{x} [m \cdot s^{-2}]$ and the pole angle angular acceleration $\ddot{\theta} [^\circ \cdot s^{-2}]$ must be calculated. Both calculations are further discussed in \cite{b4}.

As the cartpole system is controlable, an arbitrary input force $F_{cin} [N]$ can be applied, which is pointing in the same direction as the single degree of freedom of the cart. With the input force $F_{cin}$, the current system state $\mathit{S}$ and the physical constants of the cartpole, the angular acceleration $\ddot{\theta}$ of the pole can be computed as:

\begin{equation} \label{eq_cart_position_acceleration}
	\ddot{\theta} = \frac{g \cdot \sin(\theta) - \cos(\theta) \cdot \frac{F_{cin} + (m_p \cdot l_{p}) \cdot \dot{\theta}^2 \cdot \sin(\theta)}{m_p \cdot m_c}}
	{l_{p} \cdot \frac{4}{3} - \frac{m_p \cdot \cos(\theta)^2}{m_p \cdot m_c}}
\end{equation}

The cart's acceleration $\mathit{\ddot{x}}$ can be calculated using (\ref{eq_pole_angle_angular_acceleration}).

\begin{equation} \label{eq_pole_angle_angular_acceleration}
	\ddot{x} = \frac{F_{cin} + (m_p \cdot l_{p}) \cdot \dot{\theta}^2 \cdot \sin(\theta)}{m_p \cdot m_c} - \frac{(m_p \cdot l_{p}) \cdot \ddot{\theta} \cdot \cos(\theta)}{m_p \cdot m_c} 
\end{equation}

Given the previous equations, the next state $\hat{S}$ can be calculated, while dismissing any possible friction.

\subsection{Three-Tank}

\begin{figure}[htbp]
	\centerline{\includegraphics[width=9cm]{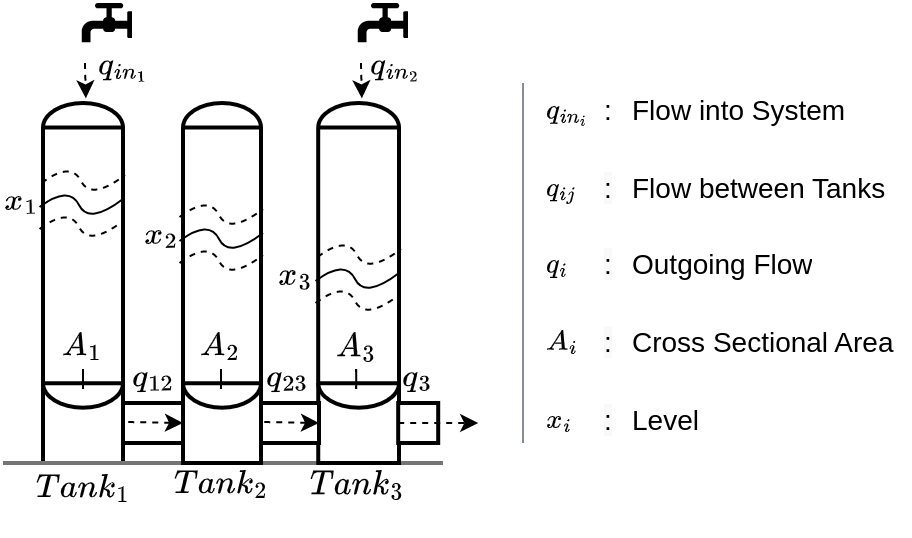}}
	\caption{Abstracted three-tank system.}
	\label{fig_three_tanks}
\end{figure}

The three-tank system consists of three connected tanks, as shown in Fig. \ref{fig_three_tanks}. $Tank_1$ and $Tank_3$ are connected to an input valve. The level $x_2 [m]$ in $Tank_2$ can only change by the flow through the pipes conneting it to $Tank_1$ ($q_{12}$) and $Tank_3$ ($q_{23}$). The outgoing flow is only possible through pipe $q_3$ connected to $Tank_3$.

For each of the tanks, the change of contained mass $\dot{M_{t}}$ is defined by any incoming flow of mass $F_{tin}$ subtracted by any outgoing flow of mass $F_{tout}$, as shown in (\ref{eq_mass_tank_in_out_overview}).

\begin{equation}
\label{eq_mass_tank_in_out_overview}
\dot{M_{t}} = F_{tin} - F_{tout}
\end{equation}

The change of mass $\mathit{\dot{M_{t}}}$ depends on the constant cross-sectional area of the tank $\mathit{A [m^2]}$, the density of the fluid $\mathit{p [kg \cdot m^{-3}]}$ and the change of the fill level $\mathit{\dot{x} [m \cdot s^{-1}]}$, as shown in (\ref{eq_tank_mass_change}).

\begin{equation}
	\label{eq_tank_mass_change}
	\dot{M_{t}} = A \cdot p \cdot \dot{x}(t),
\end{equation}

where the cross-sectional area of the tank $\mathit{A}$ and the density of the fluid $\mathit{p}$ are considered constant. In such idealized scenarios, the mass of the fluid is proportional to its volume.

The change of volume $\dot{V} [m^3 \cdot s^{-1}]$ depends on the cross-sectional area of the tank $\mathit{A}$ and the change of the level  $\dot{x} [m \cdot s^{-1}]$, which in return depends on the in- and outgoing volume-flows $q_{in}(t) [m^3 \cdot s^{-1}]$ and $q_{out}(t) [m^3 \cdot s^{-1}]$, as shown in (\ref{eq_fill_level_overview}). 

\begin{align} \label{eq_fill_level_overview}
		\dot{V} &= A \cdot \dot{x} = \sum{q_{in}(t)} - \sum{q_{out}(t)}&\\ \nonumber
		\dot{x} &= \frac{1}{A} \cdot (\sum{q_{in}(t)} - \sum{q_{out}(t)})&
\end{align}

This can be used to calculate the change of fill level $\dot{x}$ of the individual tanks. Every tank has different in- and outgoing flows, as shown in (\ref{eq_fill_level_of_individual_tanks}).

\begin{align} \label{eq_fill_level_of_individual_tanks}
		\dot{x_1} &= \frac{1}{A_1} \cdot (q_{in_1}(t) - q_{out_{12}}(t))&\\ \nonumber
		\dot{x_2} &= \frac{1}{A_2} \cdot (q_{in_{12}}(t) - q_{out_{23}}(t))&\\ \nonumber
		\dot{x_3} &= \frac{1}{A_3} \cdot (q_{in_{3}}(t) + q_{in_{23}}(t) - q_{out_{3}}(t))&
\end{align}

To calculate the in- and outgoing volume flows, the connected pipes must be simulated. As all of those are very short, their flow dynamics will not be taken into account. Furthermore, the cross section of the pipes is very small in comparison to the cross section of the tanks. This results in volume flows through the connecting pipes between the tanks that only depend on the cross section and the pipe-end pressures $\mathit{P_{A_i}}$ and $\mathit{P_{A_j}}$, which are, in this idealized construction, equal to the pressure at ground level of the tanks.

For a constant cross section, Bernoulli's equation (pressure equation) can describe the volume flow through the pipes with the cross section $\mathit{a_{ij}}$ of the pipe, an outflow constant $\mathit{\alpha_{ij}}$ with $0 <= \alpha_{ij} >= 1$, the fluid density $\mathit{p}$ and the ground pressures $\mathit{P_{A_i}}$ and $\mathit{P_{A_j}}$ of the tanks.

\begin{align} \label{eq_flow_equation}
		sgn(t) &= sign(P_{A_{j}}(t) - P_{A_{i}}(t))&\\ \nonumber
		q_{ij}(t) &= \alpha_{ij} \cdot a_{ij} \cdot sgn(t) \cdot \sqrt{\frac{2}{p} \cdot P_{A_{i}}(t) - P_{A_{j}}(t)}&
\end{align}

Furthermore, the presssures at ground level $P_A$ depend on the current level $x$ of the connected tanks. The pressures at ground level $P_A$ can be written as the force acting on the ground of the tank $F_A [N]$ regarding the cross sectional area $A$. The acting ground force $F_A$ is the product of the containing masses $M_{t}$ and the gravitational force $g$. As discussed before, the containing masses $M_{t}$ can also be expressed by a volume $V$, given the idealized system architecture. The volume $V$ is calculated by using the cross sectional area $A$ and the level of the tank $x$. Therefore, it is possible to simplify the fraction by the cross sectional area $A$.

\begin{align} \label{eq_ground_pressure_tank}
		P_{A}(t) &= \frac{F_A(t)}{A} = \frac{M_t(t) \cdot g}{A} = \frac{p \cdot V(t) \cdot g}{A}&\\\nonumber
		\frac{p \cdot V(t) \cdot g}{A} &= p \cdot \frac{A \cdot x(t) \cdot g}{A} = p \cdot g \cdot x(t)&
\end{align}

Inserting (\ref{eq_ground_pressure_tank}) in (\ref{eq_flow_equation}), the complete flow equation can be written as:

\begin{align} \label{eq_complet_flow_equation_overview}
		sgn_{ij}(t) &= sign(x_{i}(t) - x_{j}(t))&\\\nonumber
		q_{ij}(t) &= \alpha_{ij} \cdot a_{ij} \cdot sgn_{ij}(t) \cdot \sqrt{2 \cdot g \cdot | x_{i}(t) - x_{j}(t) |}&
\end{align}

The volume flow $q_3$ of pipe 3 is the outgoing flow of the system. Its pressure on the disconnected end equals the atmospheric pressure of zero. The final calculations of every volume flow of each pipe are shown in (\ref{eq_volume_flows_complete}).

\begin{align} \label{eq_volume_flows_complete}
		sgn_{ij}(t) &= sign(x_{i}(t) - x_{j}(t))&\\\nonumber
		q_{12}(t) &= \alpha_{12} \cdot a_{12} \cdot sgn_{12}(t) \cdot \sqrt{2 \cdot g \cdot |x_{1}(t) - x_{2}(t)|}&\\\nonumber
		q_{23}(t) &= \alpha_{23} \cdot a_{23} \cdot sgn_{23}(t) \cdot \sqrt{2 \cdot g \cdot |x_{2}(t) - x_{3}(t)|}&\\\nonumber
		q_{3}(t) &= \alpha_{3} \cdot a_{3} \cdot \sqrt{2 \cdot g \cdot x_{3}(t)}&
\end{align}

Inserting (\ref{eq_volume_flows_complete}) into (\ref{eq_fill_level_of_individual_tanks}), every change of each individual level $\dot{x}$ can be calculated.

\section{Data Preperation}
\label{sec:data_preperation}

The previously described simulations are used to generate training data. The data format is a list of samples, each of which represents an unique measurement session of equal length. Each time step contains either the input, the output, or a combination of both.

Random input is generated for each sample using the configuration given in table \ref{tab:model_description} under section \ref{sec:prediction_models}. The simulations produce the corresponding output, and get reset between samples.

\begin{figure}[htbp]
	\centerline{\includegraphics[width=9cm]{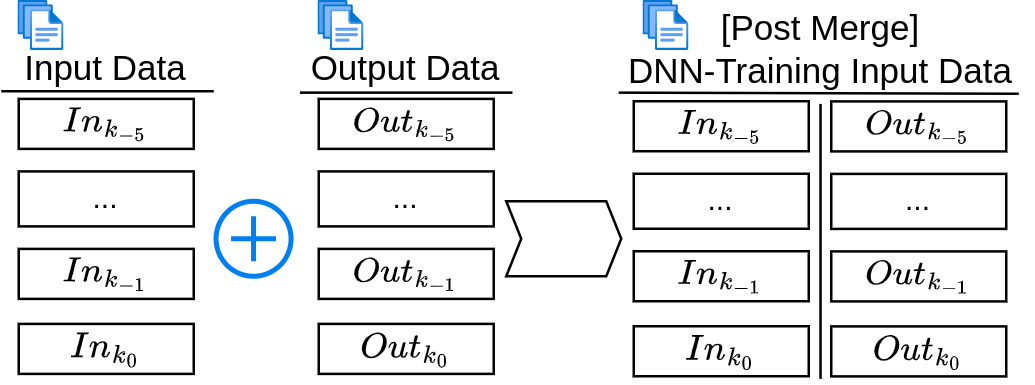}}
	\caption{Data merge example with five time steps.}
	\label{fig:data_merge}
\end{figure}

Because DNN models require a combination of input and output data for learning and predicting, the input and output data are merged, as shown in Figure \ref{fig:data_merge}.

\begin{figure}[htbp]
	\centerline{\includegraphics[width=9cm]{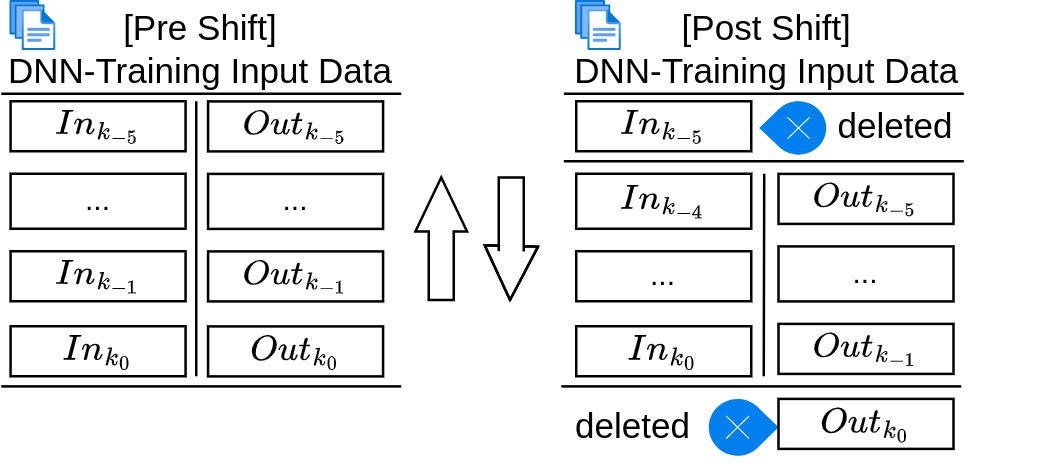}}
	\caption{Data shift example with five time steps.}
	\label{fig:data_shift}
\end{figure}

Now, one time step includes inputs and their corresponding output. But this would imply the inclusion of targets in the training data. Therefore, input and output of the merged data are shifted, so that each time step now contains the current input and the previous output, while the first input and last output of the sample are deleted, as shown in Fig. \ref{fig:data_shift}.

\begin{figure}[htbp]
	\centerline{\includegraphics[width=9cm]{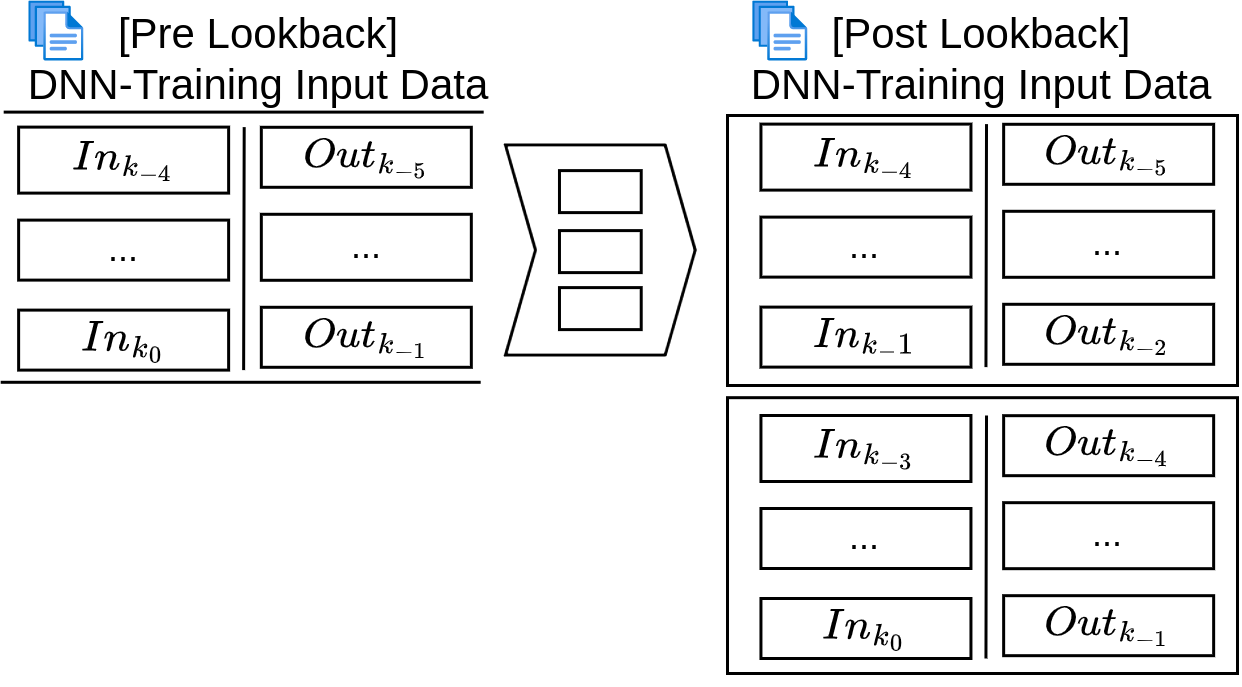}}
	\caption{Data lookback example with four time steps and a lookback of three.}
	\label{fig:data_lookback}
\end{figure}

Since the simulated systems are dynamic, the DNN needs information from past states to be able to predict the future output. This information is termed \textit{lookback} and contains the last $L$ previous values. Therefore, data are grouped into packages, each containing the current time step and its lookback, as shown in Fig. \ref{fig:data_lookback}. 

\begin{figure}[htbp]
	\centerline{\includegraphics[width=9cm]{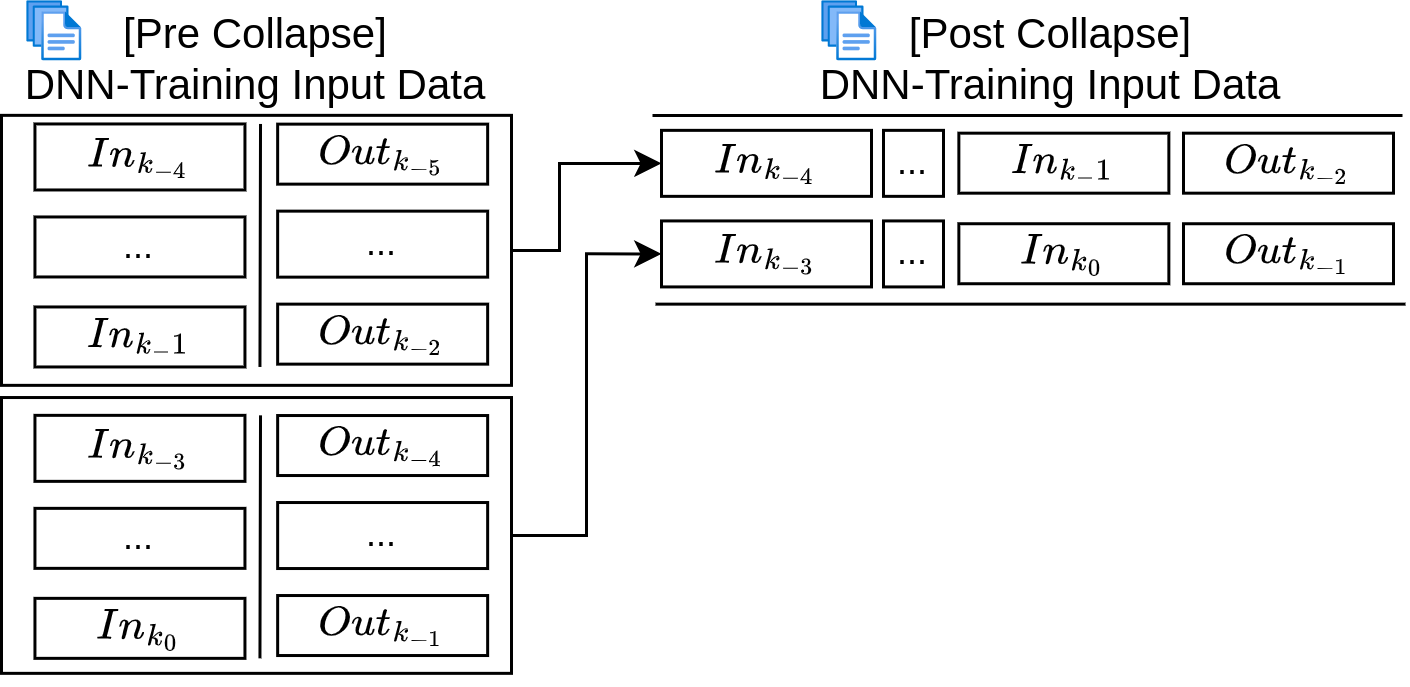}}
	\caption{Data lookback collapse example with two time steps.}
	\label{fig:data_lookback_collapse}
\end{figure}

To get the data into an understandable format for the DNN, each lookback package gets collapsed into a single tuple, as shown in Fig. \ref{fig:data_lookback_collapse}.

\begin{figure}[htbp]
	\centerline{\includegraphics[width=9cm]{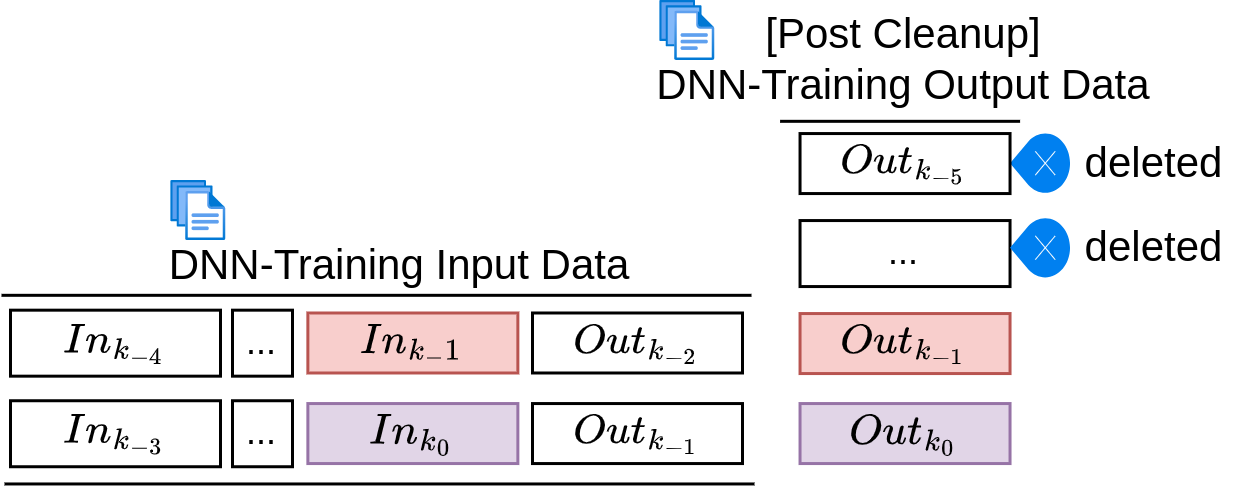}}
	\caption{Data label cleanup example for two time steps.}
	\label{fig:data_output_cleanup}
\end{figure}

Since the input data preparation removes some of the recorded time steps, the corresponding label data must also be "cleaned". Therefore, the first $L$ time steps of the output data must be removed, equal to the number of values $L$ in each lookback package. This is shown in Fig. \ref{fig:data_output_cleanup}.

\begin{figure}[htbp]
	\centerline{\includegraphics[width=9cm]{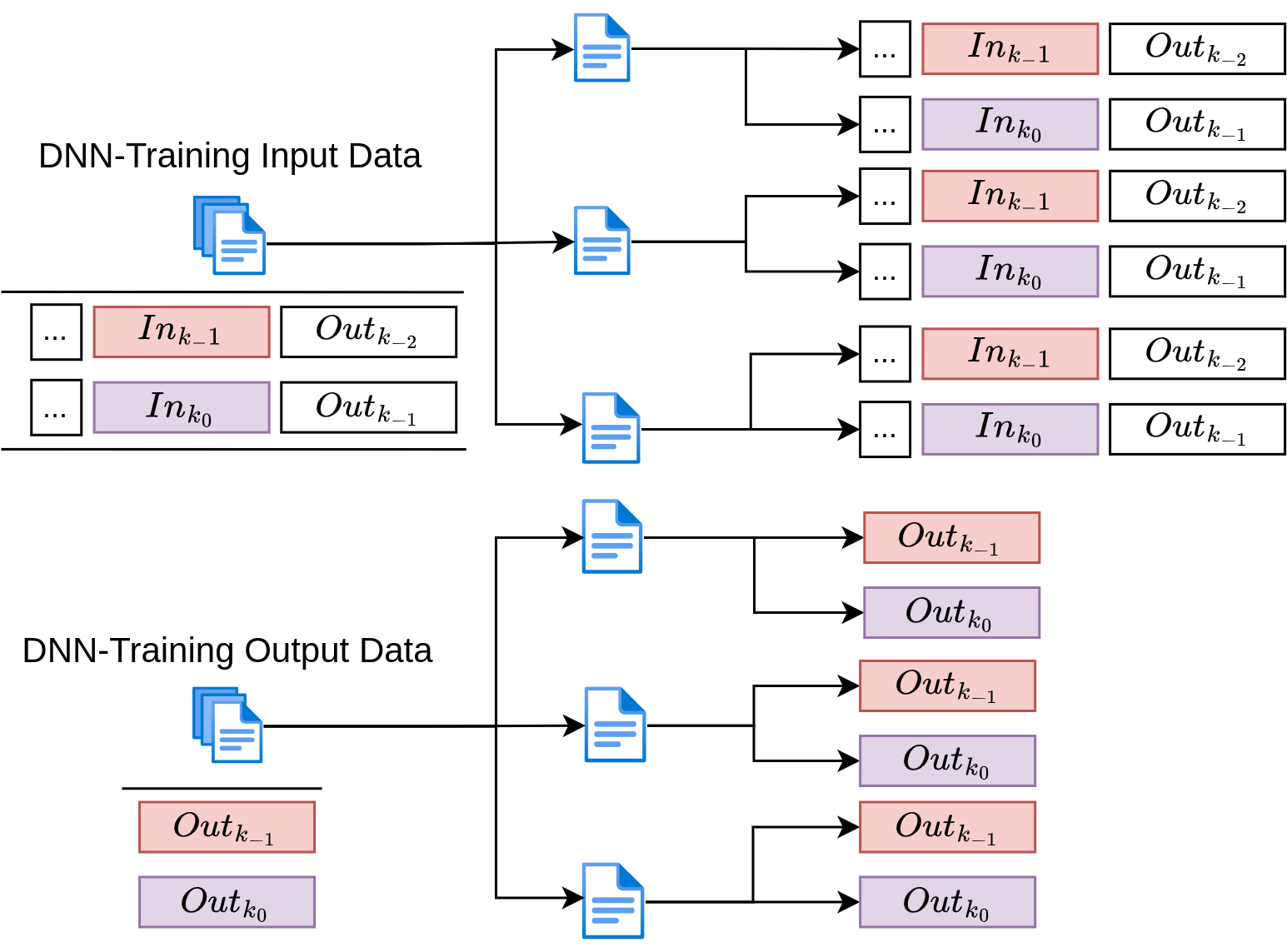}}
	\caption{Data collapse example for three samples.}
	\label{fig:data_sample_collapse}
\end{figure}

Finally, the samples of training data and targets each get collapsed into a list of time steps, as shown in Fig. \ref{fig:data_sample_collapse}.

\section{Prediction Models}
\label{sec:prediction_models}

Each MPC requires its own prediction model for the CS under consideration.

The used ML-MPC is using a single DNN architecture for each of the three problems as its PM, with the configuration shown in table \ref{tab:model_description}.

\begin{table}[htbp]
	\caption{DNN-Model Descriptions}
	\begin{center}
			\begin{tabular}{|m{0.12cm}|m{4.3cm}|m{3.2cm}|}
			\hline
				\multirow{4}{*}{\rotatebox[origin=c]{90}{\textbf{General}}} &
				\multicolumn{2}{c|}{\textbf{Activation:} Relu} \\
				\cline{2-3}
				 & \multicolumn{2}{c|}{\textbf{Loss:} mse} \\
				\cline{2-3}
				 & \multicolumn{2}{c|}{\textbf{Framework:} Keras v. 2.8.0}\\
				\cline{2-3}
				 & \multicolumn{2}{c|}{\textbf{Random Seed:} 0} \\
			\hline
				\multirow{4}{*}{\rotatebox[origin=c]{90}{\textbf{Pendulum}}} & 
				\begin{tabular}[t]{|m{1.8cm}|m{1.7cm}|} 
					\firsthline
					\multicolumn{2}{c}{\textbf{Model Summary}} \\ 
					\hline
					\textbf{Layer (type)} & \textbf{Output Shape} \\
					\hline
				 	in (InputLayer) & [(None, 8)] \\
					\hline
					dense0 (Dense) & (None, 16) \\
					\hline
					dense1 (Dense) & (None, 12) \\
					\hline
					dense2 (Dense) & (None, 8) \\
					\hline
					dense3 (Dense) & (None, 4) \\
					\hline
					output (Dense) & (None, 1) \\
					\hline
				\end{tabular} & 
				\begin{tabular}[t]{m{1.1cm} m{1.3cm}} 
					\firsthline
					\multicolumn{2}{c}{\textbf{Configuration}} \\ 
					\hline
					Samples & 1056 \\
					\hline
					Steps & 34 \\
					\hline
					Lookback & 3 \\
					\hline
					Test Split & 0.2 \\
					\hline
					Epochs & 64 \\
					\hline
					Normalize & No \\
					\hline
					\multicolumn{2}{c}{\textbf{In-/Output Ranges}}\\
					\hline
					Force & -10 - 10 \\
					\hline
					Angle & -8 - 8 \\
				\end{tabular} \\ 
			\hline
				\multirow{4}{*}{\rotatebox[origin=c]{90}{\textbf{Cartpole}}} & 
				\begin{tabular}[t]{|m{1.8cm}|m{1.7cm}|} 
					\firsthline
					\multicolumn{2}{c}{\textbf{Model Summary}} \\ 
					\hline
					\textbf{Layer (type)} & \textbf{Output Shape} \\
					\hline
					in (InputLayer) & [(None, 18)] \\
					\hline
					dense0 (Dense) & (None, 64) \\
					\hline
					dense1 (Dense) & (None, 64) \\
					\hline
					dense2 (Dense) & (None, 64) \\
					\hline
					dense3 (Dense) & (None, 64) \\
					\hline
					dense4 (Dense) & (None, 64) \\
					\hline
					dense5 (Dense) & (None, 64) \\
					\hline
					output (Dense) & (None, 2) \\
					\hline
				\end{tabular} &
				\begin{tabular}[t]{m{1.1cm} m{1.3cm}} 
					\firsthline
					\multicolumn{2}{c}{\textbf{Configuration}} \\ 
					\hline
					Samples & 7012 \\
					\hline
					Steps & 12 \\
					\hline
					Lookback & 5 \\
					\hline
					Test Split & 0.2 \\
					\hline
					Epochs & 150 \\
					\hline
					Normalize & No \\
					\hline
					\multicolumn{2}{c}{\textbf{In-/Output Ranges}}\\
					\hline
					Force & -20 - 20 \\
					\hline
					Position & -8 - 8 \\
					\hline
					Angle & -11 - 11 \\
				\end{tabular} \\ 
			\hline
				\multirow{4}{*}{\rotatebox[origin=c]{90}{\textbf{Tanks}}} & 
				\begin{tabular}[t]{|m{1.8cm}|m{1.7cm}|} 
					\firsthline
					\multicolumn{2}{c}{\textbf{Model Summary}} \\ 
					\hline
					\textbf{Layer (type)} & \textbf{Output Shape} \\
					\hline
					in (InputLayer) & [(None, 20)] \\
					\hline
					dense0 (Dense) & (None, 24) \\
					\hline
					dense1 (Dense) & (None, 24) \\
					\hline
					dense2 (Dense) & (None, 16) \\
					\hline
					dense3 (Dense) & (None, 12) \\
					\hline
					output (Dense) & (None, 3) \\
					\hline
				\end{tabular} &
				\begin{tabular}[t]{m{1.1cm} m{1.3cm}} 
					\firsthline
					\multicolumn{2}{c}{\textbf{Configuration}} \\ 
					\hline
					Samples & 806 \\
					\hline
					Steps & 24 \\
					\hline
					Lookback & 3 \\
					\hline
					Test Split & 0.2 \\
					\hline
					Epochs & 80 \\
					\hline
					Normalize & Yes \\
					\hline
					\multicolumn{2}{c}{\textbf{In-/Output Ranges}}\\
					\hline
					Inflow$_{1-2}$ & 0 - 0.0001 \\
					\hline
					Level$_{1-3}$ & 0 - 1.0 \\
				\end{tabular} \\ 
			\hline
		\end{tabular}
		\label{tab:model_description}
	\end{center}
\end{table}

The C-MPC uses the underlying simulation itself as its PM. C-MPC therefore have access to predictions that are guaranteed to produce the best possible result that the ML-MPC could ever reach.

\section{Model Predictive Control}
\label{cha:model_predictive_control}

Any MPC can be formulated as an optimization of the best next input (or: action, the language of RL) using a model of the CS for future predictions, based on an arbitrary reference value, that defines the state, the CS should reach, also often called setpoint in control science. 

Since no linearization of the discussed non-linear systems is used, the MPC includes a non-linear optimization algorithm, in particular, a custom-built genetic algorithm.

Using this approach, the ML-MPC and the C-MPC are not guaranteed to find the globally optimal action. However, since this deficiency is shared by both architectures, the comparison remains valid.

The used control loop is straightforward. The CS receives an input (action) and returns an output. Then, the MPC searches the best subsequent input based on the set of recorded outputs and the given reference value. The found input is now again applied to the CS, thus terminating one control iteration.

Within each control iteration, a non-linear optimization problem must be solved. Depending on the chosen algorithm, this takes time, energy and often does not guarantee the global minimum. Within the optimization, the PM has to make multiple future predictions, which means, a slow prediction time of the DNN or the simulation is one core issue within the non-linear optimization algorithm. Furthermore, each prediction within the optimization has to be done from the state the CS is currently in.

As the C-MPC uses the same simulation used for the CS as its PM, it can just mirror the simulations states one to one, which allows for 100\% prediction accuracy.

The ML-MPC, on the other hand, uses a DNN as its PM, where the inner states cannot be set to mirror the CS. Therefore, the current state of the CS is encoded within the input data. This is done by creating the first input state for each prediction with the past outputs of the CS and discarding any possible outputs from the DNN. This also means that the DNN can only start to predict from the correct state when enough output data of the CS is collected to create the first lookback-package for the input.

\begin{figure}[htbp]
	\centerline{\includegraphics[width=9cm]{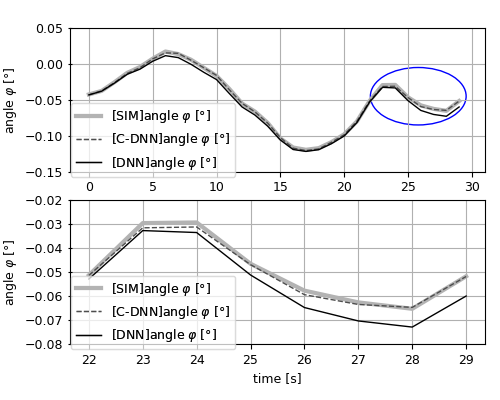}}
	\caption{Prediction session of the pendulum system with and without state correction.}
	\label{fig:correction_result}
\end{figure}

The process of encoding the states of the CS into the input data of the DNN is further called the state correction, and a DNN with corrected state is called corrected DNN (C-DNN). Shown is a comparison between a prediction session of a DNN with and without state correction in Fig. \ref{fig:correction_result}.

It is evident that the state correction process is effectivly removing the summing of prediction errors over time. Furthermore, we observe that the state of the CS can be encoded into the input data of the DNN.

With state correction, it is possible to start each future prediction within the optimization from the correct CS state. But this can only be done between each control iteration, hence the predictions within the optimization include any summed prediction errors.

However, as the state correction is effective, the DNN can be optimized to only predict the needed length for the optimization; the prediction horizon. Furthermore, any quality loss of the control cannot be descriped to possible incorrect state correction.

\section{Comparison}
\label{sec:comparison}

The final control comparison is run for each of the simulations, with its result presented in the following subsections.

\subsection{Comparison-Result: Pendulum}

\begin{figure}[htbp]
	\centerline{\includegraphics[width=9cm]{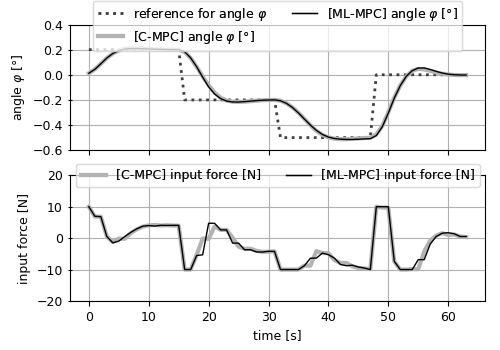}}
	\caption{ML-MPC control versus C-MPC control with three reference jumps on the pendulum simulation.}
	\label{fig:ai_nmpc_control_comparison_pendulum}
\end{figure}

The control session on the pendulum system is shown in Fig. \ref{fig:ai_nmpc_control_comparison_pendulum}.

\subsection{Comparison-Result: Cartpole}

\begin{figure}[htbp]
	\centerline{\includegraphics[width=9cm]{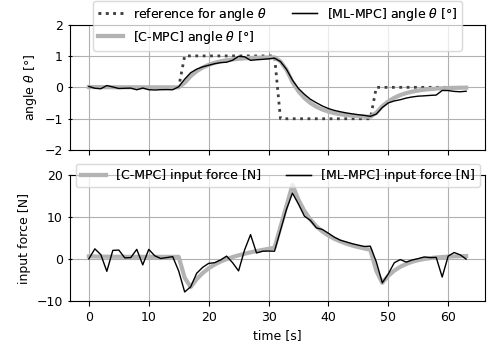}}
	\caption{ML-MPC control versus C-MPC control with three reference jumps on the cartpole simulation.}
	\label{fig:ai_nmpc_control_comparison_cartpole}
\end{figure}

The control session on the cartpole system is shown in Fig. \ref{fig:ai_nmpc_control_comparison_cartpole}.

\subsection{Comparison-Result: Three-Tank}

\begin{figure}[htbp]
	\centerline{\includegraphics[width=9cm]{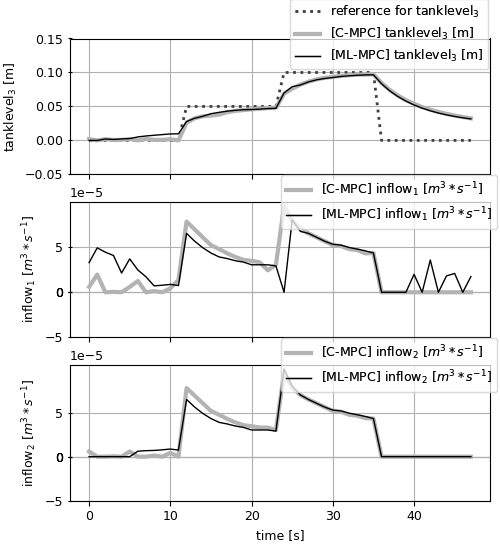}}
	\caption{ML-MPC control versus C-MPC control with three reference jumps on the three-tank simulation.}
	\label{fig:ai_nmpc_control_comparison_three_tank}
\end{figure}

The control session on the three-tank system is shown in Fig. \ref{fig:ai_nmpc_control_comparison_three_tank}.

\section{Discussion}
\label{cha:discussion}

The presented results demonstrate that the control works with varying degrees of accuracy and quality, while the C-MPC has the better result.

Since the MPCs only differ in their PMs, it is reasonable to assume that any discrepancy between their results comes from the prediction errors within the optimization.

\begin{figure}[htbp]
	\centerline{\includegraphics[width=9cm]{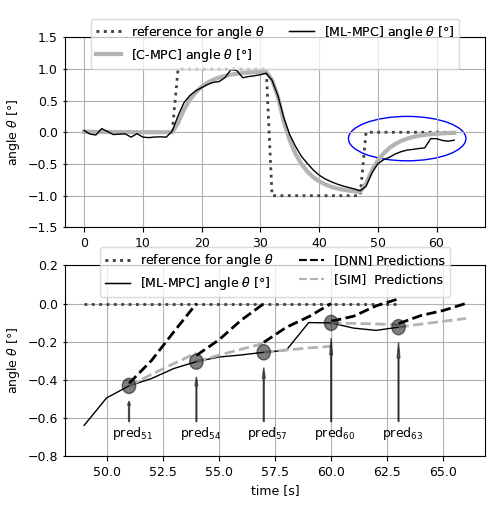}}
	\caption{Cartpole control run and zoomed in area of interest with PM predictions.}
	\label{fig:predictions_encirceled}
\end{figure}

This is shown by plotting the used predictions within the optimization and comparing them to the ideal ones, as done in Fig. \ref{fig:predictions_encirceled}.

Since the state correction was shown to be effective, the prediction errors within the control simulations are the result of the inherent prediction errors of the trained DNNs.

\section{Conclusion}
\label{cha:conclusion}

The experiments show that DNNs can be used for ML-MPCs, and that any diminished control quality regarding the C-MPC is the result of the prediction error only.

Knowing this, the ML-MPC should be improved within its separated components, where a better prediction accuracy of the ML-algorithmn directly results in a better control quality. There is no need to develop and test on ML-MPCs, as any compatible MPC-architecture can be improved without an ML-MPC in mind.

Hence, separating the issues into categories of machine learning (prediction accuracy and prediction time), control (MPC-architectures) and non-linear optimization (prediction optimization algorithm like genetic algorithm or quadratic programming) is possible and recommended, while at the same time guaranteeing an improvement of any ML-MPC.

\section*{Acknowledgment}

We thank Dr. Johann Letnev and Dr. Ivan Jursic for useful hints about the manuscript.

\end{document}